\documentclass{article}
\usepackage{spconf,amsmath,graphicx,amsfonts,amsthm,bm,xcolor,multirow, url,placeins,subfigure,flushend}

\title{ANIMATED GIF OPTIMIZATION BY ADAPTIVE COLOR LOCAL TABLE MANAGEMENT}

\name{Oliver Giudice, Dario Allegra, Francesco Guarnera,  Filippo Stanco and Sebastiano Battiato}

\address{Department of Mathematics and Computer Science, University of Catania, Catania, Italy\\
	\emph{\{giudice, allegra, fstanco, battiato\}@dmi.unict.it, francesco.guarnera@unict.it}
}

\begin{document}
	%
	\maketitle
	\begin{abstract}
		After thirty years of the GIF file format, today is becoming more popular than ever: being a great way of communication for friends and communities on Instant Messengers and Social Networks. While being so popular, the original compression method to encode GIF images have not changed a bit. On the other hand popularity means that storage saving becomes an issue for hosting platforms. In this paper a parametric optimization technique for animated GIFs will be presented. The proposed technique is based on Local Color Table selection and color remapping in order to create optimized animated GIFs while preserving the original format. The technique achieves good results in terms of byte reduction with limited or no loss of perceived color quality. Tests carried out on 1000 GIF files demonstrate the effectiveness of the proposed optimization strategy.
	\end{abstract}
	\section{Introduction}
	\label{sec:intro}
	GIF image file format, or simply GIF, is “a format that lives best on the open web, and its most important users so far have been communities of fans who make and circulate them within a participatory culture” \cite{newman2016gifs}. This is why, with the spread of social network communities, GIFs have become (again) very popular. 
	GIFs, in particular animated GIFs, were originally very common for usage on web sites through 2000s; in the last years a much more wide phenomenon raised to employ them as attachment in messages of Instant Messaging (IM) platforms like Whatsapp or Telegram. Thus a new problem arises: with respect to old way of using GIFs into web sites, now having them attached in messages of IMs makes them bigger in shear numbers, frequency, quality and length thus requiring much more storage on remote and local hosting  platforms. 
	
	GIF was designed in late 80s as a compression method, and in 30 years it has not changed at all. In this paper the animated GIF file size problem will be dealt with and tested on 1000 GIFs: the proposed method does not alter the GIF format thus having the new size-reduced files still readable from every kind of GIF reader software existing today (browsers, mobile apps, etc.). To the best of our knowledge there are no techniques that deal with this problem without altering the format itself. 
	
	The main contribution of this paper is a novel parametric size optimization technique for GIF images that still maintains image quality in terms of perceived colors and resolution. Experimental tests were carried out on 1000 proper animated GIFs which comes from the dataset {TGIF}\cite{li2016}.
	
	The remainder of this paper is organised as follows: Section \ref{sec:related} describes some historical background of GIF format and state-of-the-art optimization methods for indexed images; Section \ref{sec:proposal} deals with the proposed method with formal description; experiments will confirm the effectiveness of the technique and will be detailed in Section \ref{sec:experiments}. Finally, in Section \ref{sec:conclusion}, considerations and ideas for future work will be presented.

	\section{Historical Background and Related Work}
	\label{sec:related}
	GIF is the acronym of Graphics Interchange Format and it is a digital image file format invented in 1987 by Steve Wilhite at CompuServe\cite{GIF87}. It was originally published as 87a and then improved in 1989 \cite{GIF89a}, which is today's format. The 89a format introduced transparency and the possibility to embed, inside the same GIF, multiple images thus allowing the birth of \emph{animated GIF files}. The GIF format is an index based (palette)\cite{gonzalez2018} image representation technique that employs a loss-less compression method known as Lempel-Ziv-Welch or simply LZW published in 1984 \cite{welch1984}. Being an indexed image, a palette must be defined and the GIF format fixes its limit at 256 colors per image. As stated before, the 89a format introduced the possibility to embed multiple images into a single GIF. Thus each image can have its own 256 colors, which are specified in the file format as palette. There is a palette defined mandatory and defined for all the image that takes the name of \emph{Global Color Table} (GCT). For each other frame it is possible to optionally associate its own palette (and its own set of 256 colors) thus defining the \emph{Local Color Table} (LCT). If the LCT is not defined for a frame, the GCT will be taken into account for indexing the colors of that frame.
	
	Being an indexed image, many techniques were proposed to find an optimization solution. At first, Memon et al. \cite{memon1996ordering}  formally introduced to the community the problem of palette reordering (or re-indexing) within a framework of linear coding. The objective was to minimize the zero-order entropy of the prediction residuals of the index sequence of an image. Thus the application of computer science optimization theory to the reordering problem was carried out by many authors like \cite{battiato2004}, \cite{battiato2007}, \cite{vanhook2008}, \cite{koc2011a} and \cite{giudice2018fast}. However, many other authors tried to make better improvements by changing the original file format: the methods proposed by \cite{koc2011b} or \cite{arnavut2014}, for example, achieved great performances but by altering the file format, they produced image files not readable unless by using specific decoder.
	
	Going into specific for GIFs, the reordering processing does not produce size reduction. This is because the compression method used by GIF - namely LZW - is dictionary-based. Thus even if one of the before-mentioned techniques is employed, the optimization is obtained w.r.t. only the entropy-based compressed GIF files. Indeed, these files are not immediately readable but a de-compression step is needed before.
	
	The aim and main contribution of this paper is to propose an optimization strategy that while reducing the size of GIF files make them immediately readable by all GIF-enabled software. Commercial software, indeed, are able to convert videos to GIFs but parameters involved are only dependent to output file spatial/temporal resolution and do not deal with "GIF"-based optimization. Moreover the aim of this study was to find a method to reduce the size of files that were created without any size-related concern in a way similar to photographs acquired by mobile smartphones and compressed when shared through IMs. Finally, the proposed solution can be implemented for "online" optimization for each GIF attached on messages through IMs (client-side) or in a "parametric batch processing" for dedicated platforms like Giphy (server-side).

	\section{Proposed Method}
	\label{sec:proposal}
	In this section we describe the proposed strategy for Local Color Tables (LCT) decimation. 
	
	\subsection{Preliminary}
	We define a $H\times W$ animated GIF as a multiset of discrete functions \mbox{$\textbf{F}={\{F_i\}}_{i=1}^N$}, where $N$ is GIF's frames number. In a nutshell $F_i(\textbf{x})$ is the color of the pixel in $\textbf{x}=(x,y)$ position of the i-$th$ frame. However, GIF encoding scheme involves color-indexed images, hence we introduce a multiset of pairs $\{(I_i, T_i)\}_{i=1}^N$ such that $F_i(\textbf{x})=T_i(I_i(\textbf{x}))$ where $I_i$ is the index map and $T_i$ the color map (color table) \cite{GIF89a}, \cite{gonzalez2018}. GIF format supports up to 256 colors, then $I_i \in \mathbb{D}^{H\times W}$ where $\mathbb{D}=\{0, 1,\dots ,255\}$, whereas $T_i \in \mathbb{S}^{256}$ where $\mathbb{S}$ is the 24 bit RGB color space. Hence $T_i(k)$ is the color related to the index $k$. An animated GIF allows encoding each frame by using the Global Color Table (GCT) $G$ or a Local Color Table (LCT) $L$. Let be $\mathbf{L}=\{L_1, L_2, \dots, L_M\}$ with $M\leq N$ a set of LCTs and $G$ the GCT, then $T_i \in \mathbf{L} \cup \{G\}$. The GCT is a shared table employed by multiple frames for color indexing; on the contrary, an LCT is a table which is exclusively employed by a certain frame which presents some specific colors. Notice that thought two color tables $T_i$ and $T_j$ for $i\neq j$ could casually be equal to the same LCT in $\mathbf{L}$, this is not considered in the GIF encoding, and such LTC will be stored twice.
	
	
	The use of multiple color tables in animated GIF format was introduced to optimize the trade-off between visual quality and compression rate \cite{GIF89a}. Actually, a single color table with 256 colors might not be enough for encoding animated GIF with many frames and might lead a low visual quality. For example, in a long GIF extracted by a photo-realistic movie with millions of different colors, the quantization error introduced to pass from millions to 256 colors would badly compromise the visual quality. On the other hand, consecutive frames generally presents similar colors and, in many cases, a single table can properly cover the required color range. Since more LCTs results in higher storage space, a well-optimized animated GIF should use the minimum number of LCTs (i.e., maximize the set of frames which share the GCT) by limiting the quality drop.
	
	With this in mind, we propose a novel strategy to decrease storage space of animated GIF by reducing the set $\mathbf{L}$ of the LCTs without or with limited quality drop. We want to highlight again that this method does not change the standard encoding scheme of GIF, hence the optimized animated GIF will be still readable by any decoder. 
	
	\subsection{Proposed LCT decimation method}
	\label{sec:lctreduction}
	Our aim is to find a criterion to remove some LCTs from $\mathbf{L}$ and get the subset $\tilde{\mathbf{L}}\subset\mathbf{L}$ for the optimized GIF $\tilde{\textbf{F}}$. However, deleting a certain LCT $L_j \in \mathbf{L}$ results in replacing it with the GCT $G$, hence we propose to define a table dissimilarity measure $D(L_j,G) \in (0, 1)$ and remove $L_j$ if $D(L_j,G)\leq t$. In other words, if the LCT $L_j$ is  similar to the GCT, according to a threshold $t$, it can be replaced by the GCT with a limited quality loss. Secondly, in case the LCT related to the index map $I_i$ is removed, one has to remap the indexes in $I_i$ to make it consistent with the GCT.
	
	\subsubsection{Decimation stage}
	We define the table dissimilarity measure as follows:
	
	\begin{equation}\label{eq:distance}
	D(L_j,G)=\frac{1}{\mid G \mid} \sum_{k=0}^{\mid G \mid-1}d(L_j(k),G(p_k))
	\end{equation}
	
	\begin{equation}\label{eq:mindistance}
	p_k=\arg\min_{p \in \mathbb{D}}d(L_j(k),G(p))
	\end{equation}
	
	where $d(c_1, c_2) \in (0,1)$ is a distance between two RGB colors $c_1, c_2 \in \mathbb{S}$. Hence, the table dissimilarity measure (\ref{eq:distance}) is the average of the distances between each color $L_j(k)$ in the LCT and the most similar one $G(p_k)$ in the GCT (\ref{eq:mindistance}).
	
	\subsubsection{Remap stage}
	If $T_i$ is a LCT and $D(T_i,G)\leq t$, such $T_i$ is changed and $\tilde{T_i}=G$ is obtained. However, each index in $I_i$ points to the old color table $T_i$, then we need to remap the indexes to get the index map $\tilde{I_i}$ which correctly point to the table $\tilde{T_i}=G$. Specifically, $\tilde{I_i}$ is obtained by replacing each index $k$ in $I_i$ with the index $p_k$ according to the equation (\ref{eq:mindistance}). In a nutshell, we replace a color which occurs in a LCT with the closest one in the GCT.

	\begin{figure}[t!]
	\begin{center}
		\includegraphics[width=\columnwidth]{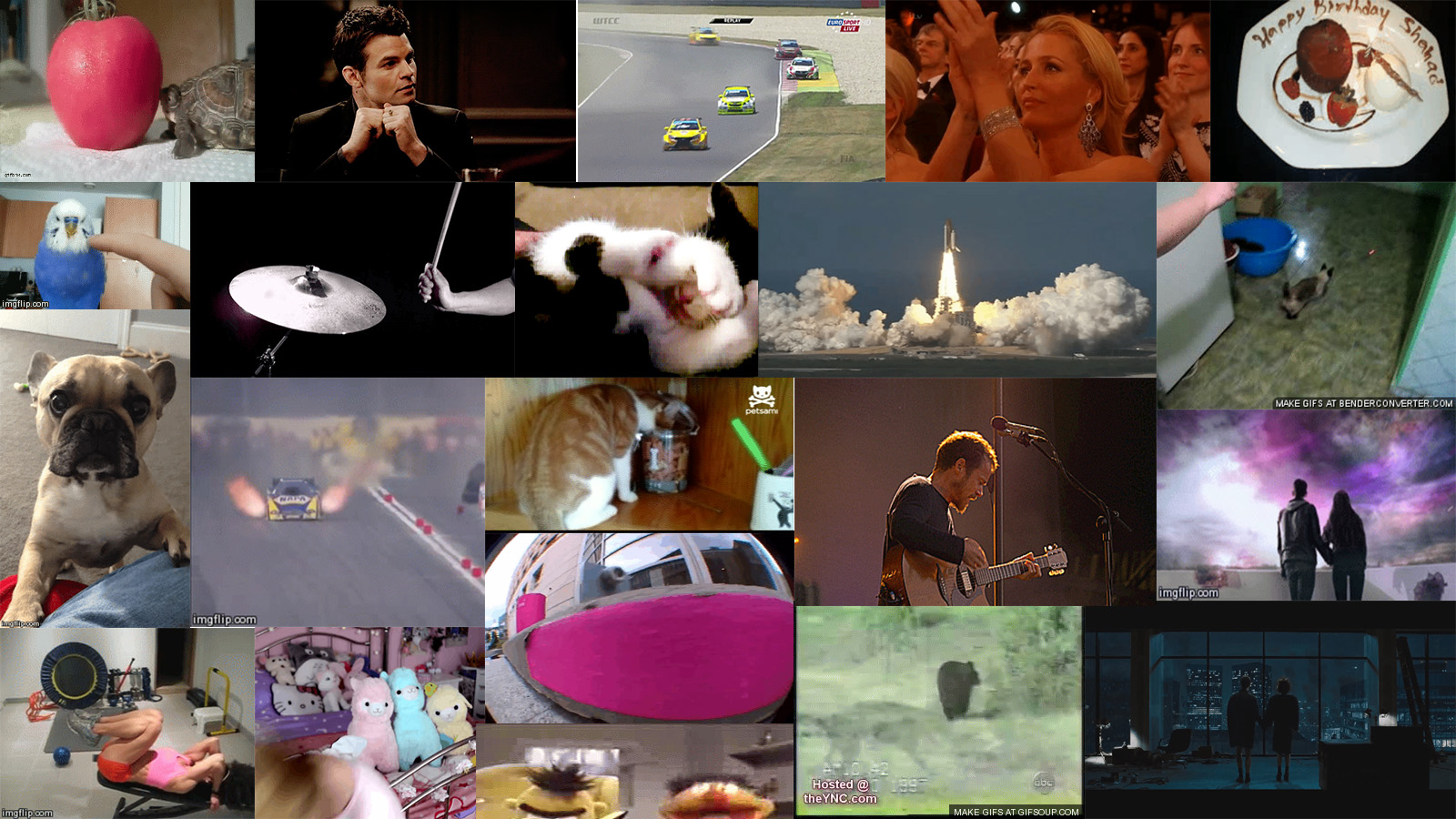}
		\caption{A sample of animated GIFs in testing dataset.}
		\label{fig:dataset}
	\end{center}
    \end{figure}

	\section{Experiments and results}
	\label{sec:experiments}
	
	To prove the validity of the proposed approach we conduct proper experiments with different threshold value $t$. To measure the color distance $d(c_1, c_2)$ between two RGB colors $c_1$ and $c_2$ we adopt a normalized version of $L^2$ distance:
	
	\begin{equation}\label{eq:colordistance}
	d(c_1,c_2)=\frac{1}{\sqrt{3}}\sqrt{\sum_{h=1}^{3}(c_1(h)-c_2(h))^2}
	\end{equation}
	
	where $c(1), c(2), c(3) \in (0,1)$ are the Red, Green and Blue components respectively. The factor $\frac{1}{\sqrt3}$ guarantees $d(c_1, c_2) \in (0, 1)$. Even if we describe this distance for RGB, it can easily adapted to any other color space.

	\subsection{Dataset}
	\label{sec:dataset}
	For testing purposes we select a subset of the dataset Tumblr GIF \cite{li2016}. TGIF includes 100k animated GIF gathered by Tumblr and 120k natural language descriptions. However, most of the GIFs in such dataset are short and have no LCTs. Hence, to properly remark the effectiveness of the proposed method, we collect 1000 animated GIF with more than 5 LCTs. In Table \ref{tab:dataset} we report dataset statistics.
	
	\begin{table}[htbp]
		\centering
		\caption{Statistics on the 1000 animated GIFs used for testing the proposed method.}
		\resizebox{\columnwidth}{!}{
			\begin{tabular}{cccccc}
				\multicolumn{1}{c}{\textbf{}} & \multicolumn{1}{c}{\textbf{Height}} & \multicolumn{1}{c}{\textbf{Width}} & \multicolumn{1}{c}{\textbf{\#frames}} & \multicolumn{1}{c}{\textbf{\#LCTs}}\\
				\hline
				Mean & 331.87 & 223.30 & 38.06 & 36.80\\
				
				Median & 320 & 206 & 34 & 33\\
				
				Max & 853 & 960 & 152 & 151\\
				
				Min & 90 & 82 & 16 & 6\\
				\hline
			\end{tabular}%
		}
		\label{tab:dataset}%
	\end{table}%

	\subsection{Evaluation metrics}
	\label{sec:metric}
	In order to evaluate the quality loss we employ the popular mean square error ($\mbox{MSE}$) and the related peak signal-to-noise ratio ($\mbox{PSNR}$); $\mbox{PSNR}=10\log_{10}(S^2/{\mbox{MSE}})$ and $S$ is the maximum value for a color component, i.e. $S=1$ in our case. Since $\mbox{MSE}$ is intended for evaluating the similarity between two images, we define it for image sequences $\textbf{F}$ and $\tilde{\textbf{F}}$ as follows:
	
	\begin{equation}\label{eq:mseavg}
	\mbox{MSE}_{avg}(\textbf{F},\tilde{\textbf{F}})=\frac{1}{N}{\sum_{h=1}^{N}{\mbox{MSE}(F_i,\tilde{F_i})}}
	\end{equation}
	
	\begin{equation}\label{eq:msemax}
	\mbox{MSE}_{max}(\textbf{F},\tilde{\textbf{F}})=\max_{i=1,\dots N}{\mbox{MSE}(F_i,\tilde{F_i})}
	\end{equation}
	
	The choice to use both the evaluation metrics, i.e. the average $\mbox{MSE}$ between GIF frames and the maximum $\mbox{MSE}$ between GIF frames, is due to the fact that the proposed strategy may keep many frames unaltered. Such frames would result in $\mbox{MSE} = 0$ (i.e., infinity $\mbox{PSNR}$) and would keep very low the error value by deceptively suggesting tremendous performance in term of quality preservation. Hence, to properly prove our method guarantees quality preservation, we also consider the maximum error $\mbox{MSE}$ for each optimized GIF. However, since $\mbox{PSNR}$ directly comes brom $\mbox{MSE}$, we report in the paper just the first one.
	
	Finally, we report the storage space saving in bit per pixel ($bpp$) at varying of the threshold $t$. It is the difference between the bit rate ($bpp$) of the original GIF and the bit rate of the optimized GIF.

	\subsection{Results}
	\label{sec:results}
\begin{figure}[t!]
	\begin{center}
		\includegraphics[width=\columnwidth]{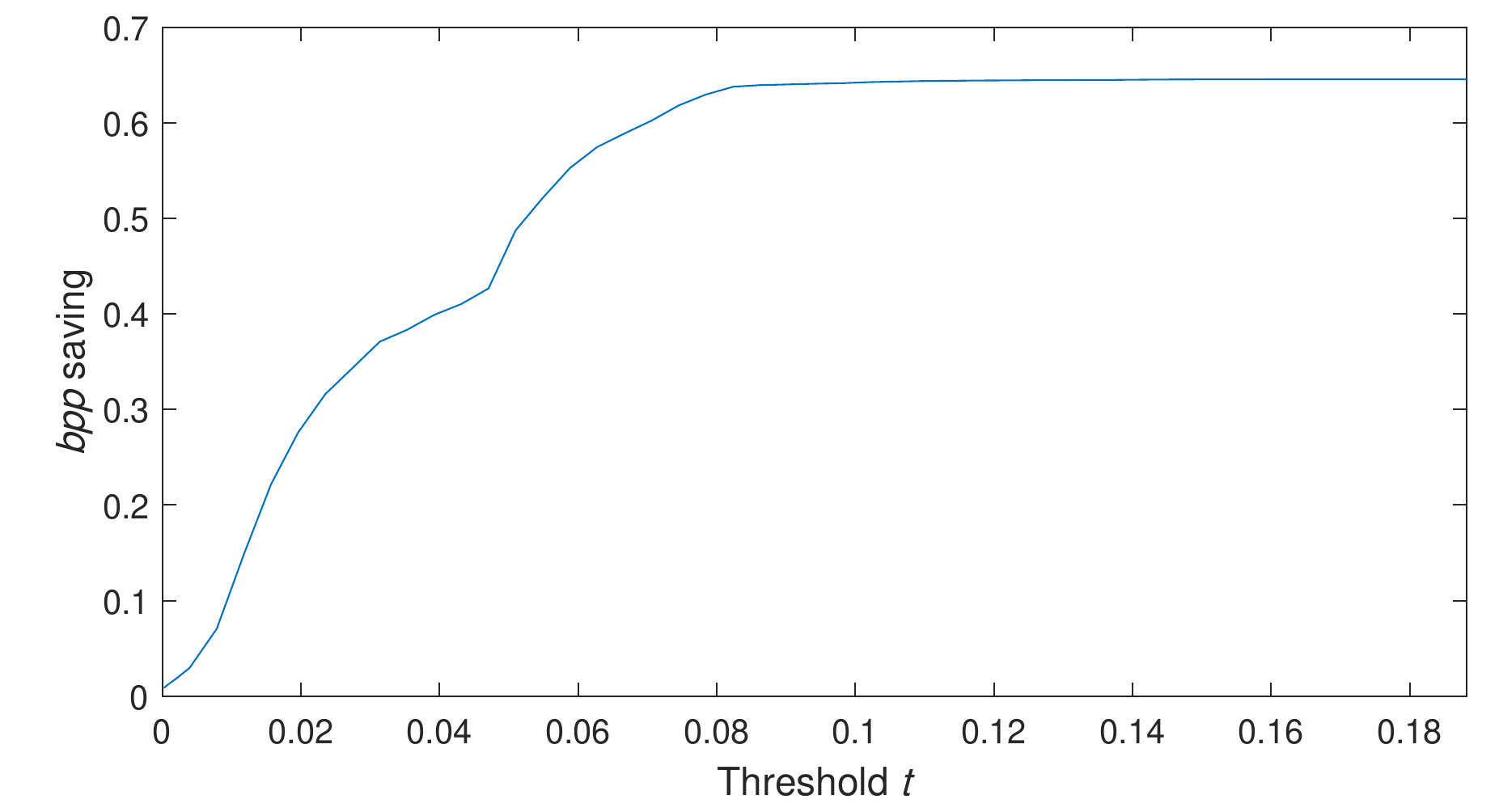}
		\caption{Storage saving ($bpp$) for different thresholds values.}
		\label{fig:tvsbpp}
	\end{center}
    \end{figure}

\begin{figure}[t!]
		\begin{center}
			\includegraphics[width=\columnwidth]{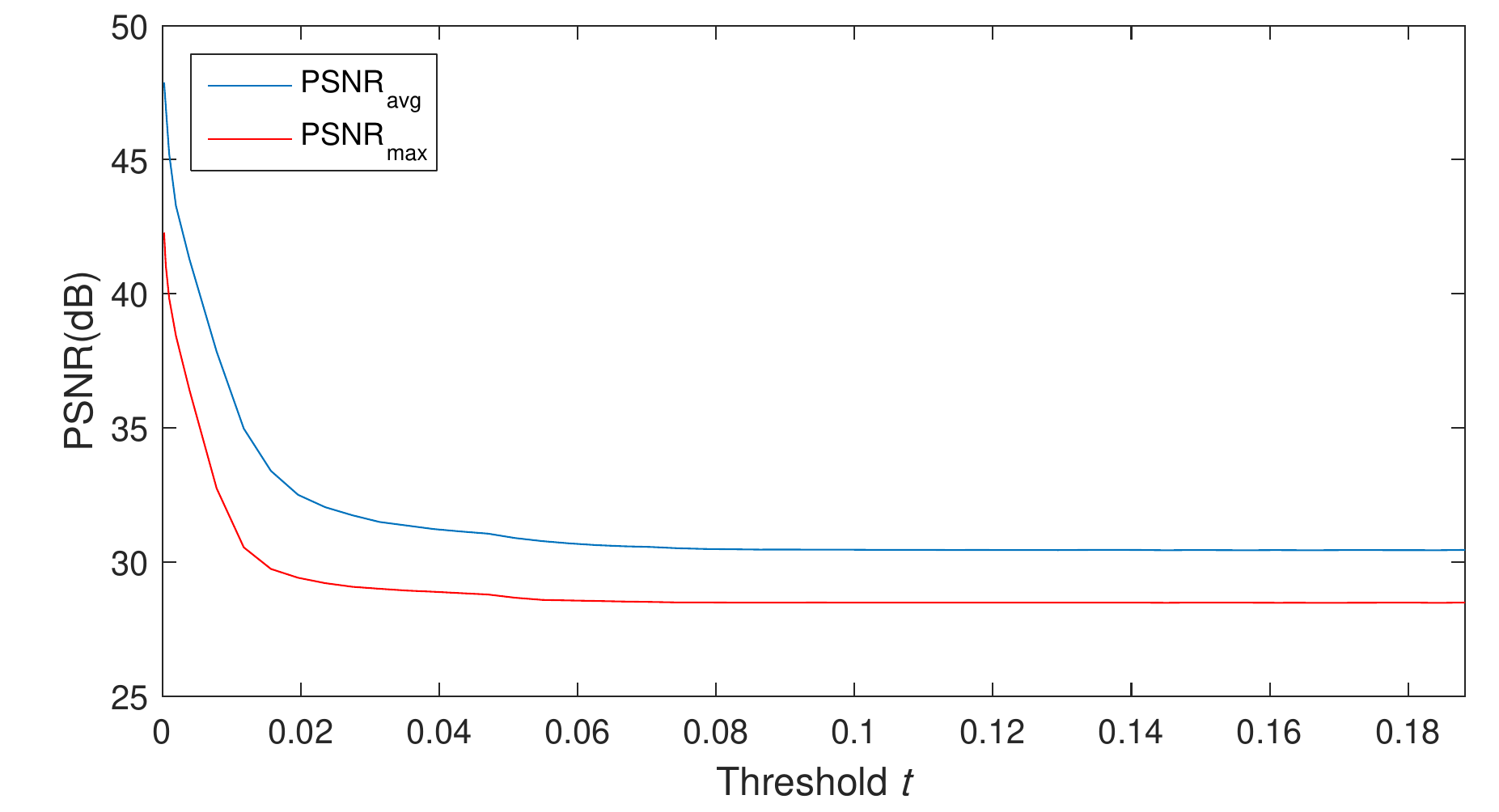}
			\caption{Quality loss ($\mbox{PSNR}_{avg}$ and $\mbox{PSNR}_{max}$) for different thresholds values.}
			\label{fig:tvspsnr}
		\end{center}
	\end{figure}

	We run the proposed method on 1000 animated GIFs with different thresholds $t$ and compare the performance in term of storage gain and quality loss. The chart in Fig. \ref{fig:tvsbpp} depict the storage saving for different threshold $t$. As expected, low thresholds values (i.e., stricter condition for removing an LCT) achieves low space gain. On the other hand, the quality loss is extremely limited and very low error (high $\mbox{PSNR}$) can be observed in Fig. \ref{fig:tvspsnr}. On the contrary, high $t$ value induces the suppression of more LCTs and results in a greater storage saving; it leads a higher, but still limited, quality drops: the minimum $\mbox{PSNR}_{avg}$ and $\mbox{PSNR}_{max}$ are $30.45$ and $28.49$ respectively for $t=0.188$.
	
\begin{figure}[t!]
		\begin{center}
			\subfigure[]{\includegraphics[height=3cm]{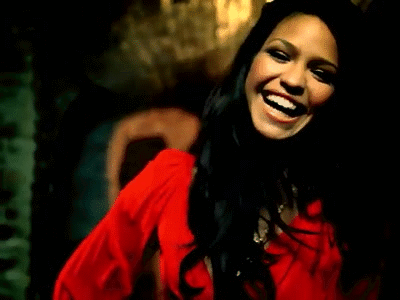}\label{fig:gif1or}}
			\subfigure[]{\includegraphics[height=3cm]{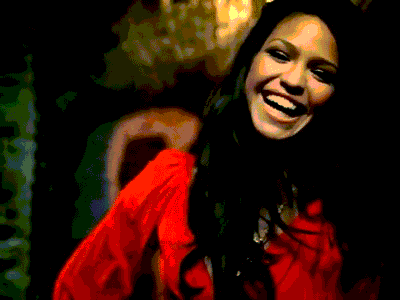}\label{fig:gif1pr}}
			\caption{(a) A frame from a testing animated GIF ($3.221$ $bpp$); (b) The same frame after GIF optimization with threshold \mbox{$t=0.125$ ($1.3222$ $bpp$)}. Although the storage gain is considerable, the quality loss can be visually perceived.}
			\label{fig:gif1}
		\end{center}
    \end{figure}

\begin{figure}[t!]
		\begin{center}
			\subfigure[]{\includegraphics[height=2.1cm]{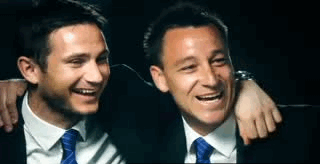}\label{fig:gif2or}}
			\subfigure[]{\includegraphics[height=2.1cm]{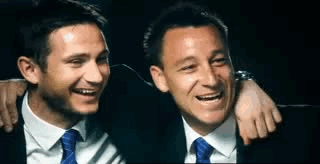}\label{fig:gif2pr}}
			\caption{(a) A frame from a testing animated GIF ($3.190$ $bpp$); (b) The same frame after GIF optimization with threshold \mbox{$t=0.020$ ($2.812$ $bpp$)}. Although the storage gain is moderate, the perceived quality is unchanged.}
			\label{fig:gif2}
		\end{center}
    \end{figure}

	For a qualitative assessment we report in Fig. \ref{fig:gif1} a case where high space saving results in higher quality loss; whereas, in Fig. \ref{fig:gif2} it is reported a case in which a moderate space saving drives no visual quality drop.
	
\begin{figure}[t!]
		\begin{center}
			\includegraphics[width=\columnwidth]{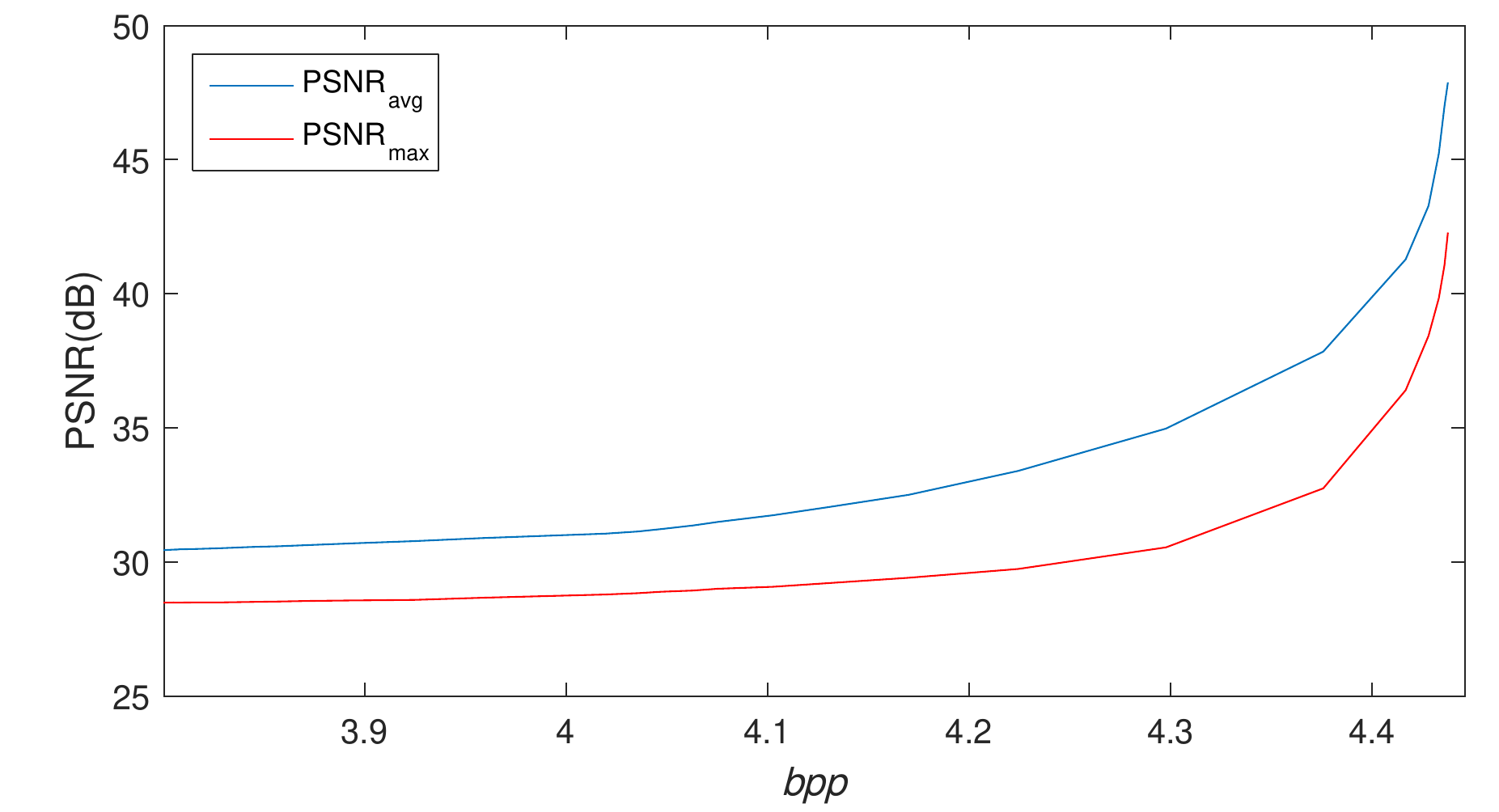}
			\caption{Distortion ($\mbox{PSNR}$) at varying of $bpp$.}
			\label{fig:bppvspsnr}
		\end{center}
	\end{figure}

	Finally, we report a distortion chart which shows the quality in term of $\mbox{PSNR}$ at varying of the average $bpp$ of the optimized GIFs for each threshold $t$ (see Fig. \ref{fig:bppvspsnr}). 
	
	\section{Conclusion}
	\label{sec:conclusion}
	GIF, while being very popular, has not changed for over thirty years. The recent increasing popularity on IMs and Social Networks asks for size optimization solutions. Commercial software aim to convert videos to GIF files but parameters related to size are strictly related to spatial/temporal resolution and do not deal with GIF.
	
	In this paper an optimization strategy for GIFs was presented which it's able to efficiently select redundant information in order to reduce size while not altering the overall perceived quality. Size reduction positively impacts on transmission time, which is a benefit for Instant Messaging platforms.
	Tests carried out on 1000 GIFs, optimizing the usage of Local and Global Color Tables, demonstrated the effectiveness of the proposed technique. The results here obtained confirm that it is possible to work on indexed animated images based on GIF format: many more improvements are still open for investigations through optimization approaches that are able to analyse all the colors of images with not only spatial but also temporal considerations.
	
	\FloatBarrier
	\bibliographystyle{IEEEbib}
	\bibliography{paper_2683}

\begin{thebibliography}{10}

\bibitem{newman2016gifs}
M.~Z. Newman,
\newblock {\em GIFs: The attainable text},
\newblock Ann Arbor, MI: Michigan Publishing, University of Michigan Library,
  2016.

\bibitem{li2016}
Y.~Li, Y.~Song, L.~Cao, J.~Tetreault, L.~Goldberg, A.~Jaimes, and J.~Luo,
\newblock ``{TGIF}: A new dataset and benchmark on animated {GIF}
  description,''
\newblock in {\em IEEE Conference on Computer Vision and Pattern Recognition},
  2016, pp. 4641--4650.

\bibitem{GIF87}
``Graphics interchange format,''
  \url{https://www.w3.org/Graphics/GIF/spec-gif87.txt},
\newblock Accessed: 2020-02-07.

\bibitem{GIF89a}
``Graphics interchange format,''
  \url{https://www.w3.org/Graphics/GIF/spec-gif89a.txt},
\newblock Accessed: 2020-02-07.

\bibitem{gonzalez2018}
R.C. Gonzalez and R.E. Woods,
\newblock {\em Digital Image Processing},
\newblock Pearson, 2018.

\bibitem{welch1984}
A.~Welch,
\newblock ``A technique for high-performance data compression,''
\newblock {\em Computer}, vol. 17, no. 6, pp. 8--19, 1984.

\bibitem{memon1996ordering}
N.~D. Memon and A.~Venkateswaran,
\newblock ``On ordering color maps for lossless predictive coding,''
\newblock {\em IEEE Transactions on Image Processing}, vol. 5, no. 11, pp.
  1522--1527, 1996.

\bibitem{battiato2004}
S.~Battiato, G.~Gallo, G.~Impoco, and F.~Stanco,
\newblock ``An efficient re-indexing algorithm for color-mapped images,''
\newblock {\em IEEE Transactions on Image Processing}, vol. 13, pp. 1419--1423,
  November 2004.

\bibitem{battiato2007}
S.~Battiato, F.~Rundo, and F.~Stanco,
\newblock ``Self organizing motor maps for color-mapped image re-indexing,''
\newblock {\em IEEE Transactions on Image Processing}, vol. 16, pp. 2905--2915,
  2007.

\bibitem{vanhook2008}
J.~{Van Hook}, F.~Sahin, and Z.~Arnavut,
\newblock ``Application of particle swarm optimization for traveling salesman
  problem to lossless compression of color palette images,''
\newblock in {\em International Conference on System of Systems Engineering},
  2008.

\bibitem{koc2011a}
B.~Koc and Z.~Arnavut,
\newblock ``Application of pseudo-distance to lossless coding of color-mapped
  images,''
\newblock in {\em International Conference on System of Systems Engineering},
  2011, pp. 220--224.

\bibitem{giudice2018fast}
O.~Giudice, D.~Allegra, F.~Stanco, G.~Grasso, and S.~Battiato,
\newblock ``A fast palette reordering technique based on gpu-optimized genetic
  algorithms,''
\newblock in {\em IEEE International Conference on Image Processing}, 2018, pp.
  1138--1142.

\bibitem{koc2011b}
B.~Koc and Z.~Arnavut,
\newblock ``Block-sorting transformations with pseudo-distance technique for
  lossless compression of color-mapped images,''
\newblock in {\em Western New York Image Processing Workshop}, 2011, pp. 1--4.

\bibitem{arnavut2014}
Z.~Arnavut, B.~Koc, and H.~Kocak,
\newblock ``Scanning paths for lossless compression of pseudo-color images,''
\newblock in {\em Western New York Image and Signal Processing Workshop}, 2014,
  pp. 11--14.

\end{thebibliography}
	
\end{document}